\documentclass[12pt, a4paper]{article}

\pdfoutput=1

\usepackage[margin=0.8in]{geometry}
\usepackage[margin=0.4in, font=small]{caption}

\usepackage{cite}
\usepackage{amsfonts}
\usepackage{amsmath,amssymb,amsthm,mathrsfs,amsfonts,mathtools,textcomp}
\usepackage{bbm}
\usepackage[utf8]{inputenc}
\usepackage[T1]{fontenc}
\usepackage{dsfont}
\usepackage{algorithm,algorithmicx,algpseudocode}
\usepackage{graphics}
\usepackage{xcolor}
\usepackage{soul}

\usepackage{tabularx}
\usepackage{booktabs}
\usepackage{hyperref}
\usepackage{url}
\usepackage{trimclip}
\usepackage{epstopdf}
\usepackage{xspace}
\usepackage{setspace}
\usepackage{float}

\def\BibTeX{{\rm B\kern-.05em{\sc i\kern-.025em b}\kern-.08em
    T\kern-.1667em\lower.7ex\hbox{E}\kern-.125emX}}

\newcounter{phase}[algorithm]
\newlength{\phaserulewidth}
\newcommand{\setphaserulewidth}{\setlength{\phaserulewidth}}
\newcommand{\phase}[1]{%
  \vspace{-1.45ex}
  \Statex\leavevmode\llap{\rule{\dimexpr\labelwidth+\labelsep}{\phaserulewidth}}\rule{\linewidth}{\phaserulewidth}
  \Statex\strut\refstepcounter{phase}\!\!\!\!\!\!\!\!\!\!\!\!\ \textit{\textbf{#1}}
  \vspace{-1.25ex}\Statex\leavevmode\llap{\rule{\dimexpr\labelwidth+\labelsep}{\phaserulewidth}}\rule{\linewidth}{\phaserulewidth}}
\makeatother
\setphaserulewidth{.3pt}

\newcommand{\featStyle}[1]		{\texttt{#1}}
\newcommand{\Ho}						{H$_0$\xspace}
\newcommand{\AUC}						{\text{AUC}}

\providecommand{\keywords}[1]
{ \small	\vspace{1.5mm}
  \noindent\textbf{Keywords ---} #1
}

\title{Revealing posturographic features associated with the risk of falling in patients with Parkinsonian syndromes via machine learning}

\author{Ioannis Bargiotas\,$^{1,2,}$\footnote{Correspondence: \texttt{ioannis.bargiotas@cmla.ens-cachan.fr}; tel.: +33 6 61 57 67 84.}, \ Argyris Kalogeratos\,$^{1}$, \ Myrto Limnios\,$^{1}$, \\Pierre\,–\,Paul Vidal\,$^{3}$$^,$$^{2}$, \ Damien Ricard\,$^{4}$, \ Nicolas Vayatis\,$^{1}$}

\date{\small
$^{1}$ CMLA, ENS Cachan, CNRS, Universit\'{e} Paris-Saclay, 94235 Cachan, France;\\
$^{2}$ COGNAG-G UMR 8257, CNRS, SSA, Universit\'{e} Paris Descartes, Paris, France;\\
$^{3}$ School of Automation, Hangzhou Dianzi University, Zhejiang, 310018, China;\\
$^{4}$ Neurology Department HIA Percy, Service de Sant\'{e} des Arm\'{e}es, Clamart, France.\\
[2ex]%
}

\begin{document}

\maketitle

\abstract{\noindent Falling in Parkinsonian syndromes (PS) is associated with postural instability and consists a common cause of disability among PS patients. Current posturographic practices record the body’s center-of-pressure displacement (statokinesigram) while the patient stands on a force platform. Statokinesigrams, after appropriate signal processing, can offer numerous posturographic features, which however challenges the efforts for valid statistics via standard univariate approaches. In this work, we present the ts-AUC, a non-parametric multivariate two-sample test, which we employ to analyze statokinesigram differences among PS patients that are fallers (PS$_{\text{F}}$) and non-fallers (PS$_{\text{NF}}$). We included 123 PS patients who were classified into PS$_{\text{F}}$ or PS$_{\text{NF}}$ based on clinical assessment and underwent simple Romberg Test (eyes open/eyes closed). We analyzed posturographic features using both multiple testing with $p$-value adjustment and the ts-AUC. While the ts-AUC showed significant difference between groups ($p$-value = 0.01), multiple testing did not show any such difference. Interestingly, significant difference between the two groups was found only using the open-eyes protocol. PS$_{\text{F}}$ showed significantly increased antero-posterior movements as well as increased posturographic area, compared to PS$_{\text{NF}}$. Our study demonstrates the superiority of the ts-AUC test compared to standard statistical tools in distinguishing PS$_{\text{F}}$ and PS$_{\text{NF}}$ in the multidimensional feature space. This result highlights more generally the fact that machine learning-based statistical tests can be seen as a natural extension of classical statistical approaches and should be considered, especially when dealing with multifactorial assessments.
}

\keywords{machine learning, multivariate two-sample statistical test, robust hypothesis testing, AUC maximization, parkinsonian syndromes, fallers, posture, balance, statokinesigram, Romberg test, force platform, Wii balance board.}

\newpage
\section{Introduction}\label{sec:intro}

Postural control is the capacity of an individual to maintain a controlled upright position. Falls have been reported as one of the major causes of injury among elderly and more importantly among patients of balance-related disorders, such as Parkinsonian syndromes (PS). It has been estimated that one third of the population over 65 years-old faces minimum one fall per year \cite{Tinetti2003}. Falls promote the decrease in mobility, problems of autonomy in daily activities (bathing, cooking, etc.), or even death \cite{Melzer2004,Tinetti2003}. Taking also into consideration the aging of many modern societies, accurate risk assessment has become a major challenge with huge socio-economic impact \cite{costOfFalls2006}. 

Force platforms are one of available acquisition tools of clinical researchers for the evaluation of postural control. Such platforms record the displacement of the center of pressure (CoP) applied by the whole body in time while the individual stands upon it and follows the clinician's instructions. These CoP trajectories, usually called statokinesigrams, have been widely used in assessing the balance disorder in healthy or PS populations. It has been shown that CoP displacement characteristics can reflect individuals' postural impairment when special acquisition protocols are followed  \cite{Melzer2004,chagdes2009multiple,Mancini2012postural}.

Clinical research often aims to find the significant differences between fall-prone individuals and others who have not yet manifested important balance impairment. Researchers usually compute several features using signal processing techniques and evaluate their usefulness relying on a variety of available univariate tests, such as the Student's t-test, Kolmogorov–Smirnov or Mann-Whitney Wilcoxon tests. However, usually in experimental works, where pre-planned hypotheses are not well-fixed, multiple univariate tests are applied consecutively in order to find the features that separate significantly the two groups. The aforementioned multiple testing scheme has been part of a well-known scientific debate \cite{feise2002multiple}, mainly criticized for the increased probability of reporting a false-positive finding. More specifically, it has been reported that for alpha level $\alpha = 0.05$, it is possible that 1 in 20 relationships may be statistically significant but not clinically meaningful \cite{feise2002multiple}. Thus, several biostatisticians recommend to disclose all the analyses that have been done, and not only the significant ones. The violation of this recommendation and the regular misuse of those tests \cite{Thiese2015} combined with the relatively small available cohorts, may lead to false conclusions and as a consequence to a significant lack of clinical consensus or at least delay in reaching it. Well-known adjustments have been proposed in order to limit the aforementioned probability of a false-positive finding (such as Bonferroni correction) but they have been reported as conservative compromises (due to the significant increase of the probability for false-negative output) \cite{feise2002multiple} that do not constitute a satisfactory solution \cite{Perneger1998}. 

Classic statistical tests are very sensitive on the size of the available dataset. The generalization of any result is not safe when only relatively small populations are available (see \cite{wood2014trap} for the high risk of making false conclusions). In order to reduce this sensitivity, machine learning algorithms assess their results using cross-validation schemes. Briefly, an algorithm trains a model that `learns' to solve the problem in a randomly selected part of the dataset (called training-set), and then tests whether it can be effective on the rest of the `unseen' data (test-set). The learning and validation process is repeated multiple times and performance metrics are averaged. In the context of multidimensional datasets with binary labels $\{-1, +1\}$, the idea of assessing the separability of two groups is based on the aforementioned learning and validation scheme. The learning process sets the criteria in order to rank the population in the test-set by means of a scoring function $s$. Those who are ranked at the top of the list will be considered to belong to the positive class \cite{Vayatis2009}. The machine learning community has recently made significant progress in this topic \cite{Clemencon2005,Gretton2012}, especially related to the design of appropriate criteria for the characterization of the ranking performance and/or meaningful extensions of the Empirical Risk Minimization (ERM) approach to this framework \cite{Agarwal2005,Cortes2004}. In a large part of these efforts, the well-known criterion of the area under the ROC curve (AUC) is considered as the gold standard for measuring the capacity of a scoring function to discriminate groups of populations \cite{Vayatis2009}. %
Briefly, in the setting of two-sample statistical testing, an algorithm `learns' the rule that maximizes the AUC between the two groups in the training-set, and then tests the applicability of this rule to the test-set during the validation process.

Unfortunately, to the best of our knowledge, these novel advancements remain largely unexploited by the parkinsonism-related community. The lack of common language and proper methodological simplifications to make the approaches easy to understand by clinical researchers are possibly the major reasons for such an observed distance. 

In postural research, simple acquisition protocols (such as the basic Romberg test) have been reported to contain inconclusive information to evaluate sufficiently the postural control of an individual \cite{palmieri2002center}. However, only recently, works proposed that a combination of multiple global features, derived from CoP trajectories using data mining techniques, might be advantageous in order to classify fallers and non-fallers. Earlier works \cite{Audiffren2016,Bargiotas2018}, showed that although none of the features alone could classify effectively elderly fallers/non-fallers (i.e.~weak classifiers), yet combining all features through non-linear multi-dimensional classification gave significant results. It is suggested that the shape of the decision surface lies indeed in a multidimensional space and should be learned using multiple features at once. As a consequence, the above findings raise reasonable questions about the ability of traditional statistical tools and testing protocols to fully reveal and exploit the existing associations.

The objective of the present study is to propose an easy-to-use-and-interpret two-sample hypothesis testing approach, in an attempt to address some the aforementioned difficulties of clinical research. Our contribution is to first propose a new variation of a multivariate two-sample test through AUC maximization, which was originally theoretically established in \cite{Vayatis2009}, and test it to a PS population which includes two groups: fallers (PS$_{\text{F}}$) and non-fallers (PS$_{\text{NF}}$). We intend to highlight the benefits that one might have by using such kind of two-sample analysis in the presence of multiple features, and demonstrate the contradicting conclusions that a traditional statistical analysis (hypothetical future clinical study) might have had compared to the proposed method.

The remainder of the article is organized as follows: Population's characteristics, acquisition protocol and analytical methodologies are presented in Sec.~\ref{sec:materials-methods}. Performance results are presented in Sec.~\ref{sec:results}. Discussion, limitations, conclusions and future perspectives are provided in Sec.~\ref{sec:discussion}.

\section{Materials and methods}\label{sec:materials-methods}

\subsection{Balance measurements and fall assessment}\label{sec:measures_assessment}
\begin{figure}[H]
\centerline{\includegraphics[scale=0.55]{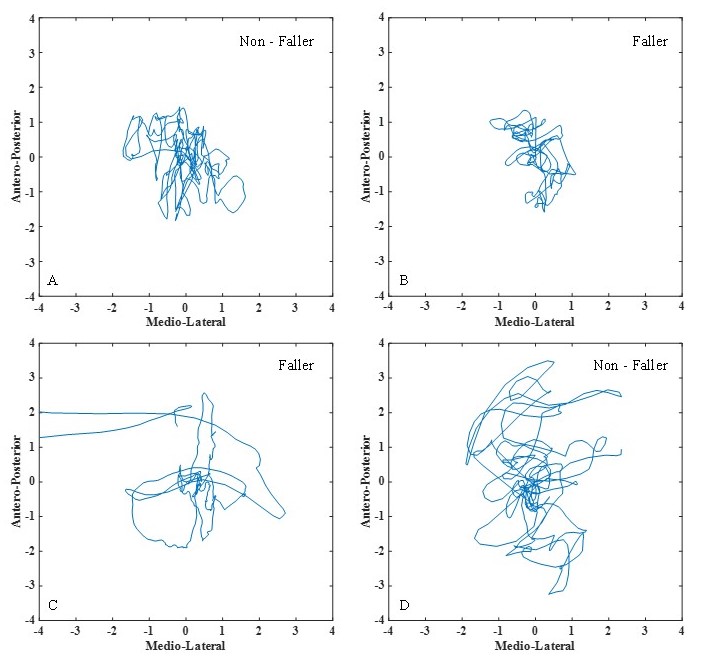}}
\caption{Examples of statokinesigrams from fallers and non-fallers. The x-axis is the medio-lateral (ML) movement and the y-axis is the antero-posterior (AP) movement of the body in centimeters (cm) during the acquisition. As it can be observed, fallers and non-fallers are not easily distinguishable by examining visually their statokinesigrams.}
\label{Statos}
\end{figure}

Our dataset comes from the Neurology department of the HIA, Percy hospital (Clamart, France), and includes 123 patients ($78.7 \pm 5.4$ years-old, Tab.~\ref{tab:demographics}) who suffered from Parkinsonian syndromes. PS patients that suffered from other comorbidities (such as vestibular and proprioceptive impairments) were not included in the study. Following the acquisition protocol, patients were asked to remove their shoes and to maintain upright position on a force platform keeping their eyes open and their arms at the side. The CoP trajectory was recorded for 25 seconds at that stance. After that, patients were asked to close their eyes maintaining their upright position. After a ten-second pause, clinical experts recorded 25 additional seconds with eyes closed (Fig.~\ref{Statos}).

\begin{table}[H]\small
\caption{Characteristics of the 123 patients included in the dataset of our experiments.}
\begin{center}
\begin{tabularx}{0.543\linewidth}{l||l|l}
\toprule
\textbf{Characteristics} & \textbf{Non-Fallers} & \textbf{Fallers}\\
\midrule
\midrule
Population & 99 & 24\\
\hline 
Age & 78.8 $\pm$ 5.3 & 78.5 $\pm$ 5.9\\ 
\hline
Gender & M:71/W:28 & M:16/W:8\\
\hline
UPDRS III total score &	23.6 $\pm$ 11.9 & 26.3 $\pm$ 11.1\\
\hline 
Disease duration & 4.7 $\pm$ 3.5 & 5.7 $\pm$ 4.2 \\
\bottomrule
\end{tabularx}
\label{tab:demographics}
\end{center}
\end{table} 

Statokinesigrams were acquired using a Wii Balance Board (WBB) (Nintendo, Kyoto, Japan), which has been found to be a suitable and convenient tool for the clinical setting \cite{clark2010validity,leach2014validating}, and the newly proposed portable package developed in our laboratory. Statokinesigram from the WBB are sent to the clinician's professional Android tablet via Bluetooth connection. Acquired signals are sent (after anonymization and encryption) to a central database for high level processing (computation of features associated to postural control and application of appropriate algorithms \cite{Audiffren2016,Bargiotas2018,Bargiotas2019balance}), and the demanded results are communicated to the clinician online. Since the  WBB records the CoP trajectories at non-stable time resolution, the acquired statokinesigrams are resampled at 25Hz using the SWARII algorithm \cite{audiffren2016preprocessing}.

In order to label the participants, a questionnaire (implemented to the Android tablet) was filled for every subject registering information about falls during the last six months prior to the examination. As in previous works \cite{Zecevic2006}, participants were labeled as fallers (PS$_{\text{F}}$) if they had come to a lower level near the ground unintentionally at least once during that period.  Twenty-four (24) patients were labeled as fallers. Any useful information about the conditions of falls were registered. The clinical trial registered at ANSM (ID RCB 2014-A00222-45) was approved by the following ethics committee/institutional review board(s): 1) Ethical Research Committees (CPP), Ile de France, Paris VI; 2) French National Agency for the Safety of Medicines and Health Products (ANSM); 3) National Commission on Informatics and Liberty (study complies with the MR-001). After information and allowing adequate time for consideration, written informed consent was obtained before participants are included in the study.

\subsection{Choice of posturographic features}\label{sec:dimension}

Our analysis included only features that were computed on the two-dimensional CoP displacement and have been previously proposed as indicators of postural impairment \cite{Baszczyk2007,Melzer2004,Muir2013}. Tab.~\ref{Table 1.} provides the names, measuring units, and descriptions (where needed) for the features that were included in the test.

\begin{table}[H]\footnotesize
\caption{Computed features derived from the CoP displacement during the acquisitions.}
\begin{center}
\begin{tabularx}{0.95\linewidth}{l|l|X}
\toprule
\textbf{Feature} & \textbf{Unit} & \textbf{Description} \\
\hline
\hline
\featStyle{RangeX} & cm & --\\
\hline
\featStyle{MaxX} & cm & Maximum medio-lateral displacement (right)\\
\hline
\featStyle{MinX} & cm & Minimum medio-lateral displacement (left) \\
\hline
\featStyle{VarianceX} & cm$^2$ & -- \\ 
\hline
\featStyle{VelocityX} & cm/s & Average instant x-axis velocity of CoP changes\\ 
\hline
\featStyle{AccelerationX} & cm/s$^2$ & Average instant x-axis acceleration of CoP changes \\ 
\hline
\featStyle{F95X} & Hz & Frequency below which 95\% of the x-axis CoP trajectory’s energy lies\\ 
\hline
\featStyle{RangeY} & cm & --\\ 
\hline
\featStyle{MaxY} & cm & Maximum antero-posterior displacement (front)\\ 
\hline
\featStyle{MinY} & cm & Minimum antero-posterior displacement (back)\\
\hline
\featStyle{VarianceY} & cm$^2$ & --\\
\hline
\featStyle{VelocityY} & cm/s & Average instant y-axis velocity of CoP changes \\
\hline
\featStyle{AccelerationY} & cm/s$^2$ & Average instant y-axis acceleration of 
CoP changes \\
\hline
\featStyle{F95Y} & Hz & Frequency below which 95\% of the y-axis CoP trajectory’s energy lays) \\
\hline
\featStyle{DistC} & cm & Instant distance from the center of the trajectory \\
\hline
\featStyle{EllArea} & cm$^2$ & Confidence ellipse area that covers the 95\% of the trajectory's points \\
\featStyle{AngularDeviation} & degrees &	Average of the angle of deviation\\
\bottomrule
\end{tabularx}
\label{Table 1.}
\end{center}
\end{table}

\subsection{Two-sample test through AUC optimization (ts-AUC)}\label{sec:ts-AUC}

\begin{figure}[t]
\centerline{\includegraphics[scale=0.60, viewport=0 180 740 535, clip]{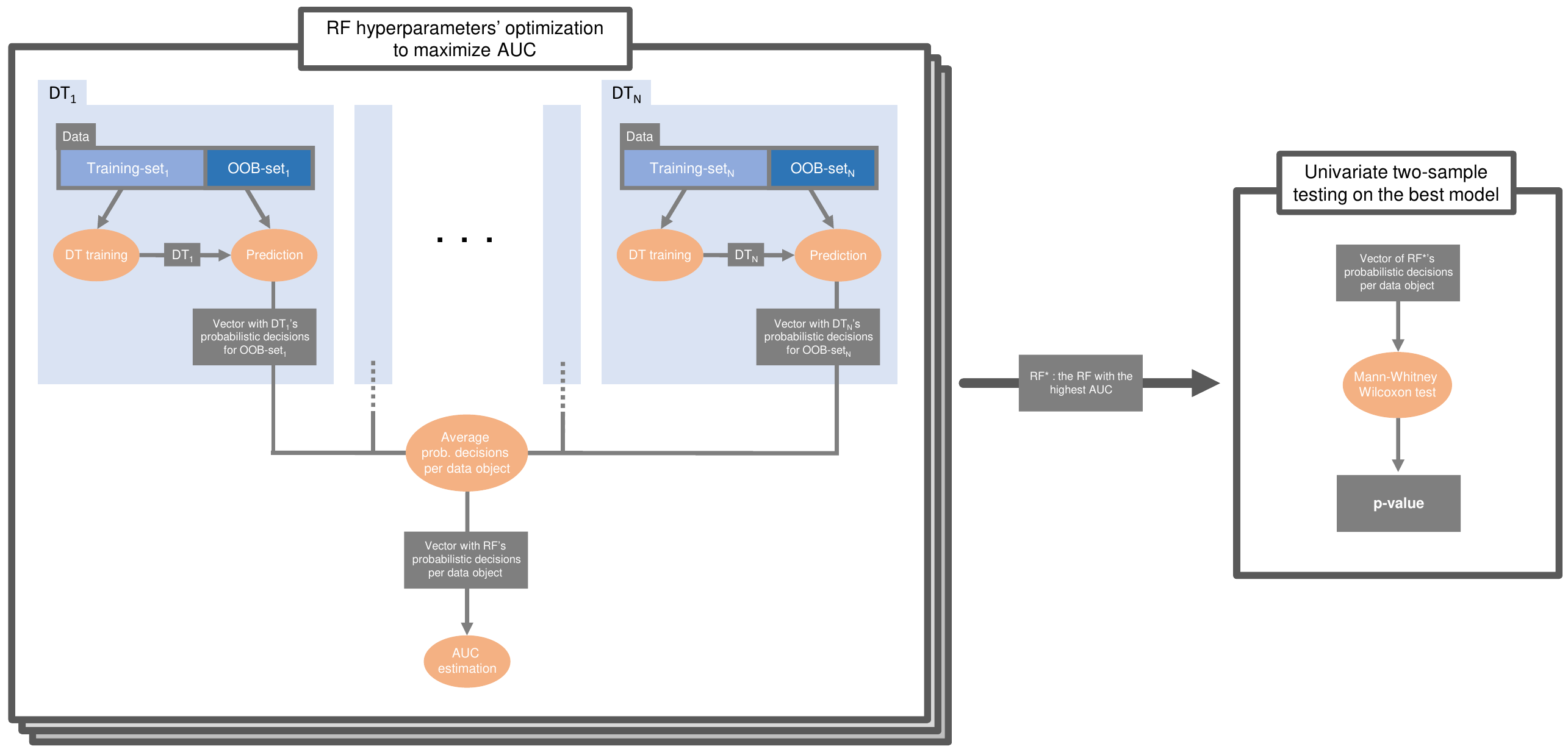}}
\vspace{1mm}
\caption{Scheme of the ts-AUC algorithm. In order to find the AUC$^*$ (maximal AUC), a number of Random Forests (RFs). For the RF$^*$ with the best AUC$^*$, the univariate Mann-Whitney Wilcoxon non-parametric two-sample test is applied on the average posterior probability values of the whole population.}
\label{Diagram}
\end{figure}

Although the proposed algorithm originate from \cite{Vayatis2009}, herein we present some algorithmic and cross-validation modifications. In the current work we use a bootstrap aggregation classification, in particular a random forest (RF) \cite{Breiman2001} that comprises several decision trees (DTs). Therefore, in the development of each DT, only a part of the whole dataset does participate (in-bag) while the other part is left out (out-of-bag, or OOB). Consequently, the OOB subset can be used as test-set for the the particular DT. In our approach, instead of the originally proposed testing method based on data splitting, we used the predictions of the OOB population \cite{Breiman1996}. The number of DTs was large enough ($N = 200$) compared to the actual population. The individuals can be selected in different OOB sets more than once. Every time an individual is part of an OOB set, the corresponding DT outputs the probability for him/her being a PS$_{\text{F}}$ or a PS$_{\text{NF}}$. This is computed as the fraction of individuals of the positive class (fallers) in the tree leaf where he/she reaches. Thus, his/her final score is given by the average of the posterior probabilities over the trees he/she was part of the OOB set (see Fig.~\ref{Diagram}). Averaged posterior probabilities ($P$) of the positive class (fallers) are used in order to compute the Mann-Whitney $U$-test statistic, denoted by $U$. The empirical AUC for the chosen hyper-parameters is given by $\frac{U}{N_{\text{F}} \cdot N_{\text{NF}}}$. Briefly, the null hypothesis, \Ho, and the alternative one, $\text{H}_\text{1}$, are expressed as follows:

\begin{equation}\label{eq:hypothesis_test}
``\,\text{H}_{\text{0}} : \AUC^* = \frac{1}{2}\text{\,''}\quad\text{\it{versus}}\quad``\,\text{H}_{\text{1}} : \AUC^* > \frac{1}{2}\text{\,''}.
\end{equation}
The OOB percentage was fixed to 36.8$\%$ of the included population. Searching the empirical $\AUC^*$  (maximal $\AUC$), the hyper-parameters that are optimized are the leaf-size $LS$ and the number of features to be used per tree $M$. We avoided a greedy approach using a Bayesian optimization process where only relatively shallow ($7<LS<20$) and simple ($M<9$) DTs were allowed to be tested. The averaged posterior probabilities of the Star Model, where $\AUC = \AUC^*$, are used to compute the scoring function (and the $p$-value) through a univariate Mann-Whitney Wilcoxon (MWW from now on) test on the whole available dataset (see Alg.~\ref{alg:ts-AUC} and Fig.~\ref{Diagram}).

\begin{figure}[t]
\centering
\begin{minipage}{0.87\linewidth}
\begin{algorithm}[H]\small
\caption{The proposed ts-AUC statistical test.}
{\bf Input:}  $X$ and $Y$ are the points' coordinates of the trajectory (statokinesigram); \\ $LS$, $OOB$, $M$ are vectors with the required hyper-parameters.

{\bf Output:} $AUC^*$, $RF^*$, $P^*$, $p$-\emph{value}$^*$.
\vspace{-4pt}
\begin{spacing}{1.05}
\begin{algorithmic}[1]
\phase{\ \ Step 1: Exploration of the space of hyperparameters}\vspace{-0.7mm}%
\For{$i \in LS$}
\For{$j \in M$}
\State $RF = \text{RandForest}\,(X, Y, LS_i, M_j)$
\State $P = \text{OOBpredict}\,(RF_{i,j})$
\State $U = \text{Mann\_Whitney\_Utest\_Statistic}\,(P)$
\State $AUC_{i,j} = \text{AUCestimation}\,(U, Y)$
\EndFor
\EndFor
\phase{\ \ Step 2: Choose the best model and apply MWW}\vspace{-0.7mm}%
\State $(i^*,j^*) = \arg\max_{i\in LS, j \in M}AUC_{i,j}$
\State $AUC^* = AUC_{i^*,j^*}$ 
\State $RF^* = \text{RandForest}\,(X, Y, LS_{i^*}, M_{j^*})$
\State $P^* = \text{OOBpredict}\,(RF^*)$
\State $p$-\emph{value}$^* = \text{MWW}\,(P^*,Y)$
\end{algorithmic}
\end{spacing}
\label{alg:ts-AUC}
\end{algorithm}
\end{minipage}
\end{figure}

\subsection{Out-of-bag feature importance}

Additionally, the proposed algorithmic modifications allow also the assessment of the importance of each feature to the ts-AUC final decision. We estimated also the out-of-bag feature importance by permutation. Briefly, the more important a feature is, the higher its influence (i.e.~the increase) would be to the model's error after feature's random permutation at the OOB subset. The permutation of a non-influential feature will have minimum, or no effect at all, on the model’s error. Having $D$ features in the dataset and $T$ trees in the RF model, the influence of feature $j \in \{1,...,D\}$ is computed as: 
\begin{equation}\label{eq:feature_importance}
I_j = \frac{\ d_j\ }{\ \sigma_j\ },
\end{equation}
where $d_j$ is the average change of model error after the permutation of feature $j$, and $\sigma_j$ is the standard deviation of the above change. Important to explain that every feature $j$ participates only to the training of a subset of the trees of the RF. Therefore, $d_j$ and $\sigma_j$ are derived by those trees in which the feature $j$ was selected to participate in their training.

Since our objective is to enhance interpretability of results, our feature importance analysis aims to identify all the important features, even those which are redundant or colinear, rather than finding a parsimonious set of important features. Hence, we followed the additional procedure proposed in \cite{genuer2010variable} especially for interpretation purposes. Briefly, we computed the AUC of the OOB ($\AUC_{\text{OOB}}$) of RFs starting from the most important feature, and adding progressively all the others in descending importance order. The best model is the smallest model (less features) with an $\AUC_{\text{OOB}}$ higher than the maximum $\AUC_{\text{OOB}}$ reduced by its empirical standard deviation (based on 20 runs). 

\subsection{Experimental settings}\label{sec:experimental_setting}
We compare the results obtained by the proposed ts-AUC with the  Maximum Mean Discrepancy test (MMD-test) \cite{Gretton2012}, which is a well-established  multivariate test and state-of-the-art in terms of performance. The MMD measures the maximum difference between the mean of two data samples, in the space of probability measures of a Reproducing Kernel Hilbert Space (RKHS). Practically, this test is the unbiased squared MMD statistic. It has been proven to be highly efficient and easy to use (an available package with kernel optimization is provided in \cite{sutherland2016generative}).

In addition, we compare the results of ts-AUC with standard statistical testing approaches which are usually used in clinical studies. We checked the $p$-values of all 17 features (i.e.~$D = 17$) with the labels \{`faller'/`non-faller'\} using the non-parametric Mann-Whitney Wilcoxon test. Typically, clinicians would report those features which were found statistically significant (e.g.~with $p\text{-value} < \alpha = 0.05$) and any interesting non-significant finding.

In order to prevent the increase of the false positive probability due to the large number of tested hypotheses, $p$-value adjustment procedures are applied. We use the Bonferroni correction, which is the most widely used $p$-value adjustment in biomedical research. Moreover, after taking into account the criticism that Bonferroni has received \cite{Perneger1998}, we also apply alternative approaches such as Holm-Bonferroni \cite{Holm1979} and Sidak corrections \cite{Sidak1967}. 

Finally, we assess the effect of population size to the final result by performing the following two additional experiments: 
\begin{itemize}
\item[1)] We progressively decrease, uniformly at random, the population size by a step of 10\% (95\% to 35\%). 
\item[2)] We progressively reduce, uniformly at random, the number of PS$_{\text{NF}}$ by a step of 10\% (95\% to 35\%). 
\end{itemize}
At every step, all analyses run 12 times and the percentages of significant results were compared (see Fig.~\ref{NonFallerDecreaseO} and Fig.~\ref{SampleDecreaseO}).

\section{Results}\label{sec:results}

The presented ts-AUC test was applied using the features derived from statokinesigrams from Eyes-Open and Eyes-Closed acquisitions. Tab.~\ref{Table 2.} contains the obtained $p$-values for the two groups by the application of the ts-AUC and MMD tests. Both these tests agreed that the features derived by statokinesigrams of Eyes-Open significantly separated PS$_{\text{F}}$ from PS$_{\text{NF}}$, contrary to those from Eyes-Closed that did not show a significant result (Tab.~\ref{Table 2.}). Therefore, we will henceforth continue by presenting detailed analysis only for Eyes-Open features. 

\begin{table}[b]\small
\caption{The $p$-values obtained by the application of the ts-AUC and MMD tests on the features extracted from Eyes-Open and Eyes-Closed statokinesigrams. Features derived by Eyes-Closed statokinesigrams did not show a statistically significant result neither using ts-AUC nor MMD test. Therefore the study did not proceed to further analysis of these statokinesigrams. The statistically significant results are indicated by `\,*\,'.}
\begin{center}
\begin{tabularx}{0.51\linewidth}{l||l|l}
\toprule
\textbf{Data type} & \textbf{MMD result} & \textbf{ts-AUC result}\\
\midrule
\midrule
Eyes-Open & \Ho rejected * & $p\text{-value} < 0.01$ *\\
\hline
Eyes-Closed	& \Ho not rejected & $p\text{-value} > 0.05$\ \\
\bottomrule
\end{tabularx}
\label{Table 2.}
\end{center}
\end{table}

\begin{figure}[t]
\centerline{\includegraphics[scale=0.75]{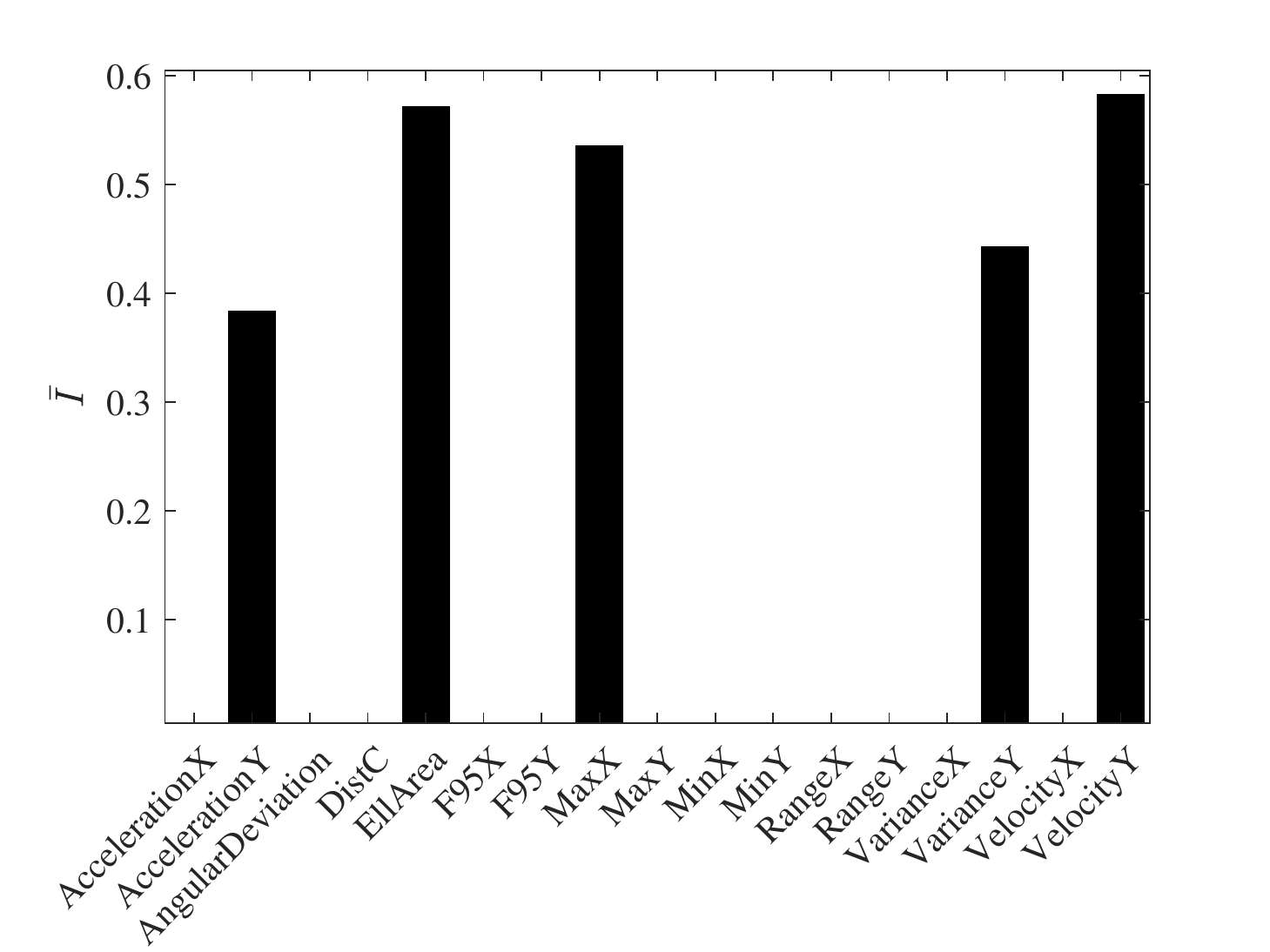}}
\caption{The importance of features as estimated by applying the approach of \cite{genuer2010variable} using the hyperparameters that produced the RF$^*$. }
\label{PredictorImportance}
\end{figure}

\begin{figure}[!b]
\centerline{\includegraphics[scale=0.8, viewport=70 50 410 300, clip]{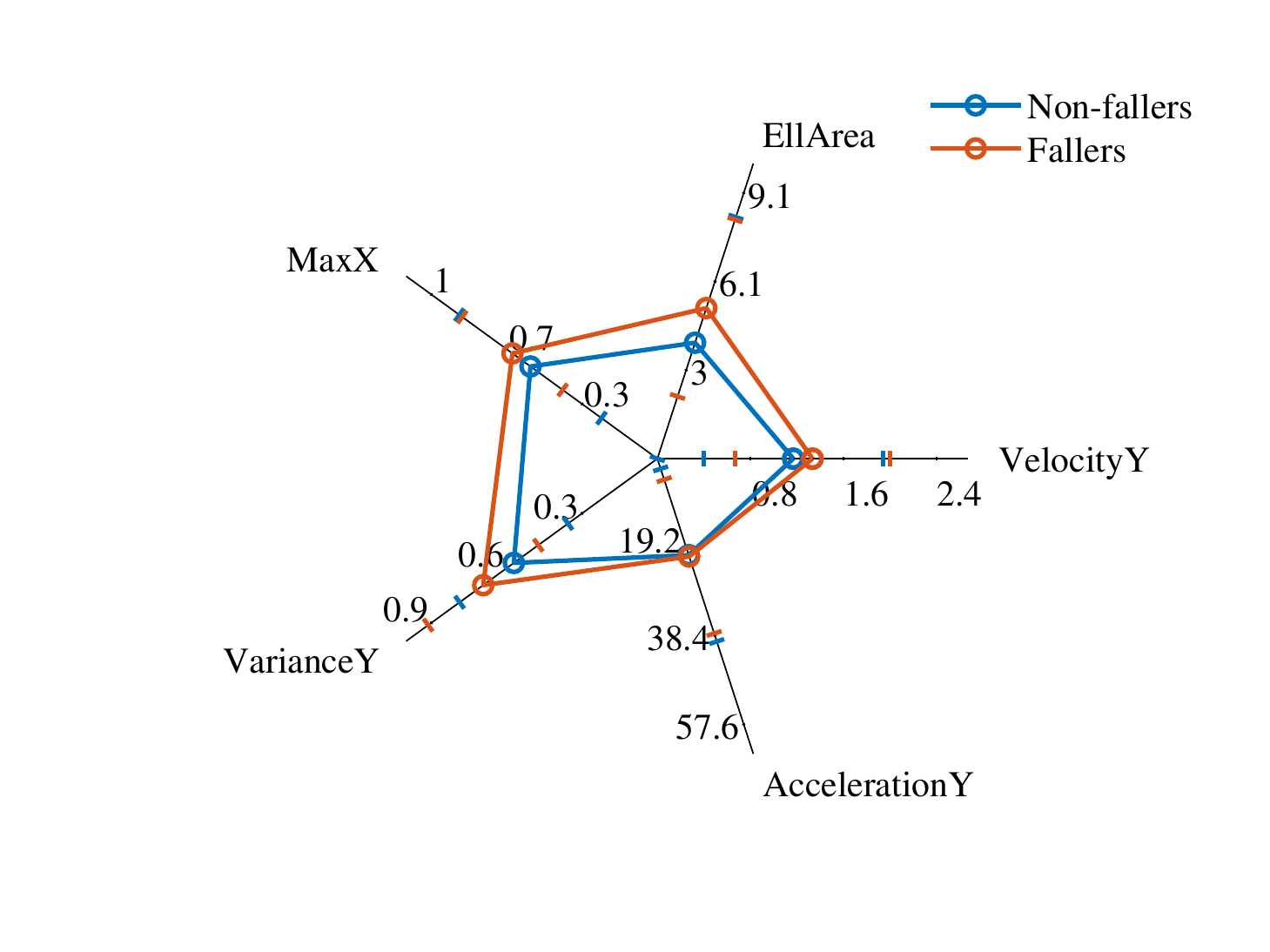}}
\caption{Radar chart comparing fallers and non-fallers based on the mean (o) $\pm$ standard deviation (-) of the most important features of our analysis. All six features are positively correlated with low postural control, which justifies the meaningfulness of inspecting the area of the curves in this chart. The profile of the two groups is significantly different.}
\label{RadarChart}
\end{figure}

The most influential features were found to be the \featStyle{VelocityY}, \featStyle{VarianceY},  \featStyle{AccelerationY}, \featStyle{EllArea} (Confidence Ellipse area), and \featStyle{MaxX} (see in Fig.~\ref{PredictorImportance} their relative importance and in Fig.~\ref{RadarChart} their mean $\pm$ standard deviation per group).
Tab.~\ref{Table 3.} indicates those features that showed $p\text{-value} < 0.05$ and the decisions regarding statistical significance obtained after applying each of the three employed corrections. Interestingly, although the \featStyle{AccelerationY} did not show statistical significance after the MWW application ($p\text{-value} > 0.05$), it was found as one of the influential features by the ts-AUC test. According to Tab.~\ref{Table 3.}, using the results from the three corrections with level $\alpha = 0.05$, none of the features would reject the \Ho of two-sample MWW test.  

\begin{table}[t]\footnotesize
\caption{Significant results of a univariate two-sample Mann-Whitney Wilcoxon (MWW) test, and the $\alpha$ levels of significance after Bonferroni, Holm-Bonferroni, and Sidak corrections. Every $p$-value presented in the MWW column is compared with the corresponding level of significance. After the corrections, $p$-values derived by MWW were found to be always greater than the corresponding level of significance. Therefore, none of the features can reject the null hypothesis of equal medians at the default 5\% significance level.}
\vspace{-1mm}
\begin{center}
\begin{tabularx}{0.9\linewidth}{X||X||X|X|X}
\toprule
 & & \multicolumn{3}{c} {\textbf{Levels of significance after correction}}\\
\textbf{Feature} & $p$-value of MWW & Bonferroni & Holm-Bonferroni  & Sidak \\
\midrule
\midrule
\featStyle{EllArea} & 0.0045 & 0.0029 & 0.0029 & 0.003 \\ 
\hline
\featStyle{VarianceY} &	0.006 & 0.0029&	0.0033 & 0.003 \\
\hline 
\featStyle{MaxY} & 0.006 & 0.0029 & 0.0036 & 0.003 \\
\hline
\featStyle{DistC} &	0.007 &	0.0029&	0.0031 & 0.003 \\
\hline
\featStyle{RangeY} & 0.008 & 0.0029 &	0.0038 &  0.003 \\
\hline
\featStyle{VelocityY} &	0.009 & 0.0029 &	0.0071 & 0.003 \\
\hline
\featStyle{MaxX} &	0.03 & 0.0029 &	0.0045 & 0.003 \\
\hline 
\featStyle{RangeX} & 0.04 & 0.0029 & 0.005 & 0.003 \\
\hline
\featStyle{VarianceX} &	0.04 & 0.0029 &	0.0042 & 0.003 \\
\hline
\featStyle{MinY} &	0.04 & 0.0029 &	0.0063 & 0.003 \\
\bottomrule
\end{tabularx}
\label{Table 3.}
\end{center}
\end{table}

\subsection{Population size}

\begin{figure}[!b]
\centerline{\includegraphics[scale=0.45]{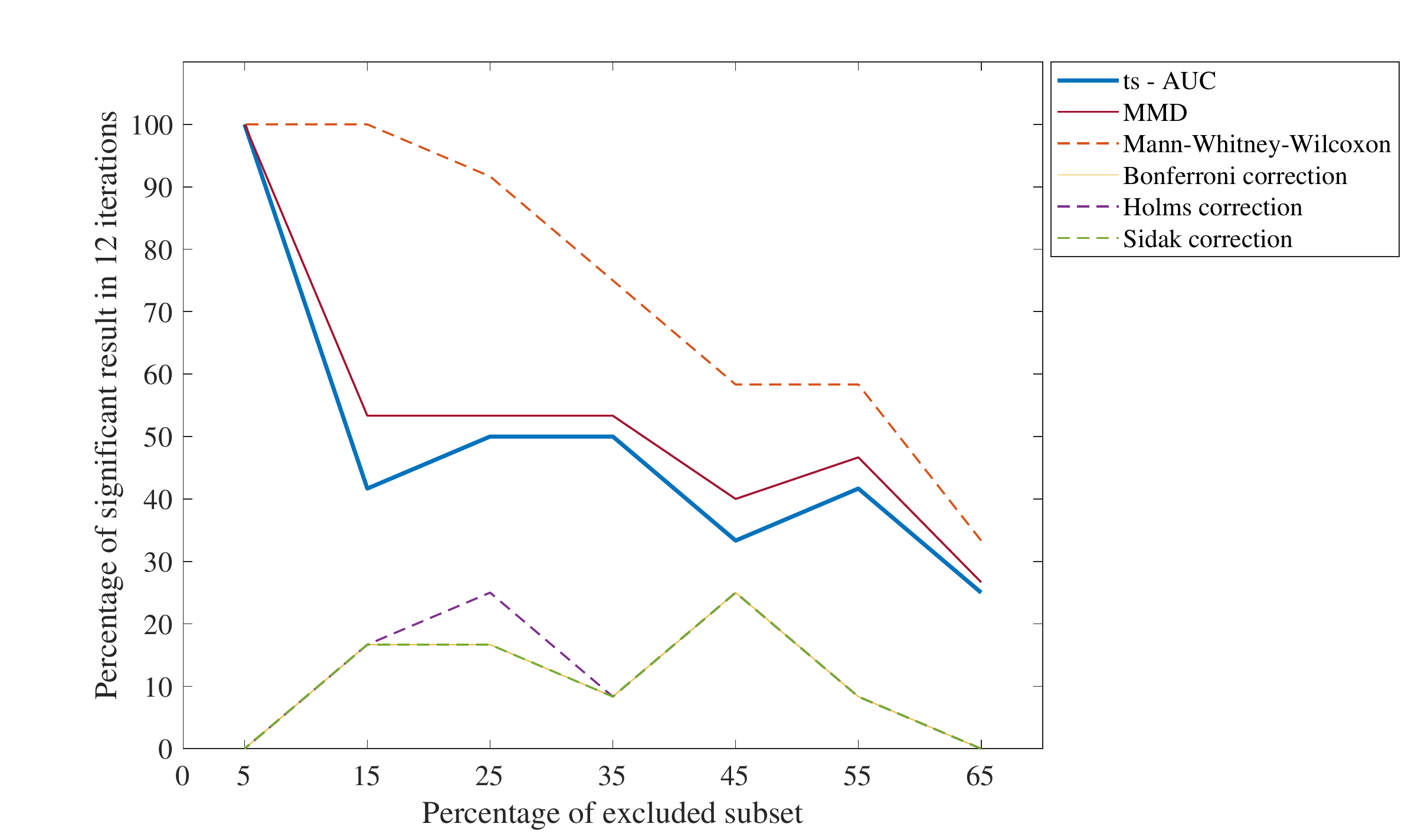}}
\vspace{-0.5mm}
\caption{The average performance of two-sample testing approaches with smaller population. The dataset size was progressively decreased by a step of 10$\%$. The included subset of each step was selected uniformly at random 12 times and the tests run in every iteration. We observe that ts-AUC and MMD have almost the same performance. Decreasing the population leads to lower chance of distinguishing the two groups. On the other hand, all the two-sample corrections present significantly lower performance.}
\label{SampleDecreaseO}
\end{figure}

As expected, the decrease of population size had an important effect to the performance of all tests. Both ts-AUC and MMD test showed similar behavior with the progressive decrease of population size. Specifically, the number of times that the fallers and non-fallers were found statistically different was gradually decreased. After 55\% of population size decrease, the two groups were found significantly different in less than 50\% of the cases (Fig.~\ref{SampleDecreaseO}). Univariate testing with MWW followed a similar decrease. Multiple testing showed that the the groups cannot be considered as statistically different (almost always).

\begin{figure}[t]
\centerline{\includegraphics[scale=0.45]{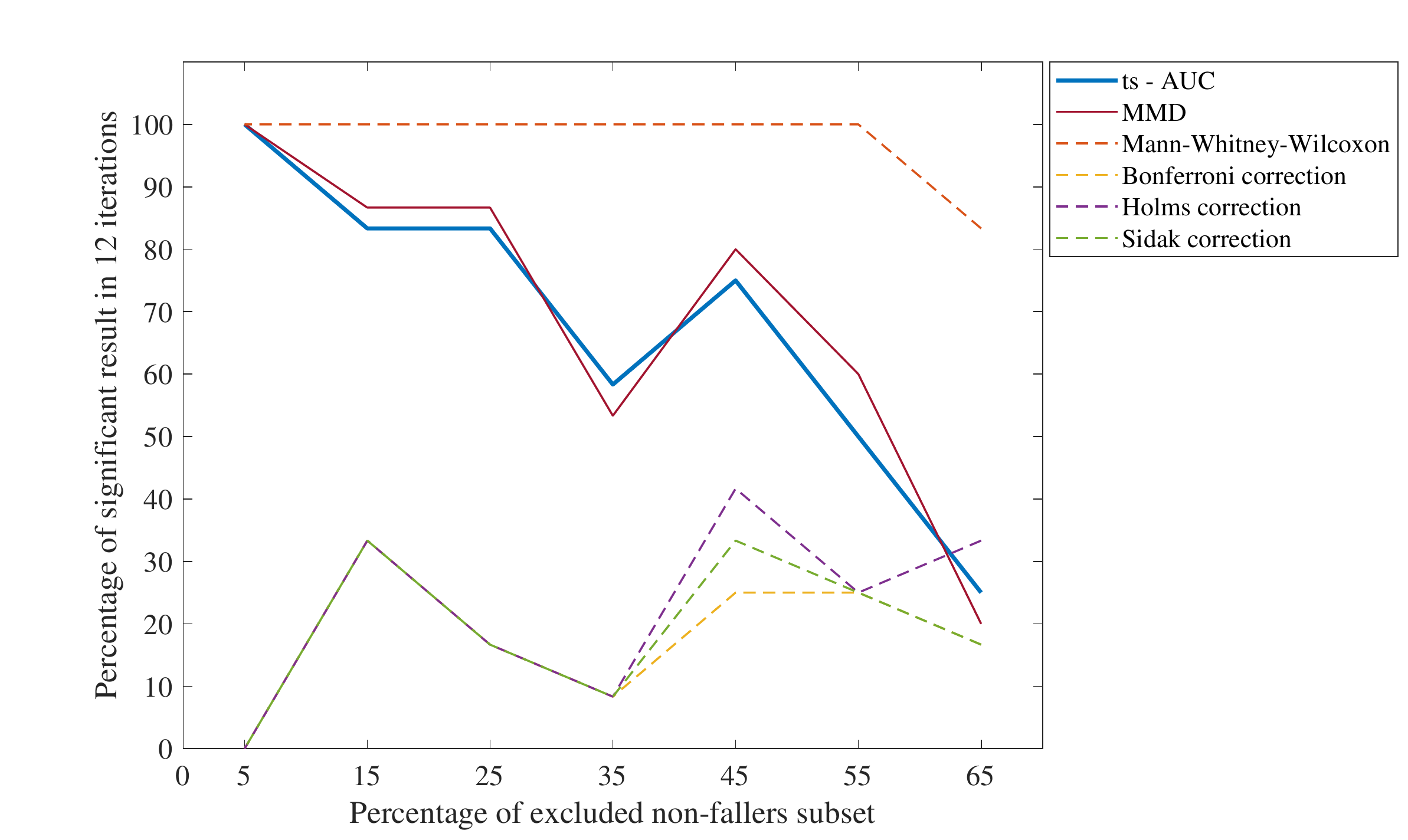}}
\caption{The average performance of two-sample testing approaches with smaller non-faller population. The non-fallers were progressively excluded, by a step of 10\%, in order to balance the size of the two groups without excluding fallers. The included subset of each step was selected uniformly at random 12 times, all fallers were included, and the tests run in every iteration. We observe that ts-AUC and MMD have almost equal performance. Decreasing the non-faller population leads to lower chance of distinguishing the two groups. On the other hand, all the two-sample corrections present significantly lower performance.}
\label{NonFallerDecreaseO}
\end{figure}

\vspace{0.5mm}
Regarding Fig.~\ref{NonFallerDecreaseO}, that shows the important role of the size proportion among the groups,   the performance of ts-AUC, MMD, and multiple testing were comparable to those from Fig.~\ref{SampleDecreaseO} (uniform decrease of the population size). However, ts-AUC and MMD exhibit a less abrupt decrease of performance. On the other hand, the gradual balancing of the sizes of the two groups, through the exclusion of non-fallers, seems to have a minor effect on the univariate MWW testing. 

\section{Discussion}\label{sec:discussion}

The objective of this study was to introduce an easy, interpretable, and intuitive multivariate two-sample testing strategy.
The particular interest of this study was to highlight the beneficial effect that this approach can have in clinical research, and particularly in the research of postural control in PS patiens.
Using the proposed statistical testing approach, it was shown that: a)~Different profiles between fallers and non-fallers were observed only for Eyes-Open protocol; %
b)~The fall-prone PS patients have significantly different statokinesigram profile during quiet standing from those who are non-fallers, contrary to the classic multiple testing approach which did not agree with such a result; %
c)~The novel multivariate two-sample testing approach (ts-AUC) showed equal performance with the state-of-the-art Maximum Mean Discrepancy (MMD) test, with the additional element of providing feature importance assessment. d) The \featStyle{VelocityY}, \featStyle{VarianceY}, \featStyle{AccelerationY}, \featStyle{EllArea} (Confidence Ellipse area), and \featStyle{MaxX}, appeared to be the most important features for distinguishing fallers and non-fallers. 

\subsection{Comparison between multivariate and multiple testing}

One of the main results of this article is that the proposed multivariate two-sample test, the ts-AUC, and the standard statistics (usually used in clinical studies), when both applied to the dataset of PS patients lead to contradictory conclusions. The multivariate approach found fallers' and non-fallers’ statokinesigram characteristics significantly different, while traditional statistics did not confirm this result. The disagreement of the traditional approach seems to be linked to the relative conservatism of the traditional $p$-value correction strategies (increase of probability of false-negative findings) \cite{feise2002multiple,Perneger1998}.
 
Researchers can always perform multiple univariate tests and not apply correction strategies (see univariate MWW results in Tab.~\ref{Table 3.}, Fig.~\ref{SampleDecreaseO}, and Fig.~\ref{NonFallerDecreaseO}), and take the risk of having a false-positive finding. However, when modest evidence is found in relatively small populations after multiple testing, then the aforementioned false-positive probability is significantly high. The level of that risk may be controlled when some criteria are met (see \cite{feise2002multiple}) considering the quality of the study, the quality of the dataset and the clinical strength of pre-set hypotheses. In exploratory studies though, some of the $p$-values around 0.05, whichever side they may lie on, would definitely be considered as ``interesting hints'', whereas concluding without thoughtful consideration from such findings should be generally avoided \cite{wood2014trap}. The multivariate and cross-validated approaches can decrease the aforementioned uncertainty. The proposed ts-AUC test has interesting and convenient properties: it is a test which is easy to implement and interpret, while it can be also applied to other similar multidimensional datasets.

\subsection{Posturographic profiles - PS$_{\text{F}}$ versus PS$_{\text{NF}}$}

The features included in our analysis have been used by clinical researchers in the past. Most of them were proposed as indicators of balance impairment at least once in the clinical literature (indicative references \cite{Baszczyk2007,Mancini2012,Melzer2004,Muir2013}). We deliberately avoided any feature engineering or transformation process, not only because that goes beyond the scope of this study, but also because we intended to focus particularly on the merits of the newly proposed approach.

Interestingly, only the Eyes-Open acquisition allowed to significantly distinguish fallers from non-fallers in a population of PS patients. This result seems slight contradictory since PS patients exhibit increased dependency on visual sensing \cite{rinalduzzi2015balance}.  By exploiting the advantage of the ts-AUC test that provides automatically the importance assessment of features, we found that medio-lateral movement played also a role in faller/non-faller separation of PS patients (see Fig.~\ref{PredictorImportance} and Fig.~\ref{RadarChart}). The medio-lateral movement has been reported as the most discriminative element between PS patients and age-matched controls \cite{Mancini2012postural} and seems that play a role in distinguishing fallers and non-fallers PS patients. However, the key-difference between fallers and non-fallers was spotted in antero-posterior movement. \featStyle{VelocityY}, \featStyle{VarianceY}, and \featStyle{AccelerationY}, which may carry overlapping information, were found among the most influential features for the separation fallers/non-fallers separation. The aforementioned result is in line with previous works that reported increased antero-posterior movement of PS patients in quiet-standing conditions with eyes open \cite{kerr2010predictors,matinolli2007postural,latt2009clinical}. 
Although many PS patients with low postural control did not manifest large posturographic areas, the confidence ellipse area (\featStyle{EllArea}) was found significantly larger in fallers compared to non-fallers (Fig.~\ref{RadarChart}). However, the \featStyle{EllArea} value of non-fallers was highly dispersed. Therefore larger fallers cohorts are needed in order to draw safer conclusions. The confidence ellipse area is recommended to be always considered together with antero-posterior features such as variance and velocity, in order to perform more accurate postural control evaluations. 

\subsection{Algorithmic aspects}

The choice of using the OOB observations as cross-validation method has two basic advantages: 1)~provides faster results in the AUC maximization process, and 2)~allows the final MWW test to be applied once to the whole dataset, which is more intuitive for clinicians. In cases where the population size is sufficiently large and the hypothesis of similar distributions between train and test-sets is not violated, it is expected that more classic methods such train-test split (as originally proposed in \cite{Vayatis2009}) would have given the same result (or even better; OOB prediction error results have been reported as slightly overestimated \cite{janitza2018overestimation}). However, clinical datasets are usually limited in size and the aforementioned assumption about the same distribution is not always fully guaranteed. In these cases, multiple train-test splits seem more appropriate whereas they would significantly increase the testing process. OOB observations can be seen as an internal multiple train-test split (one per tree-learner) of the RF (each observation’s prediction is predicted by less than $N$ trees) but with the nice intuition that the final two-sample MWW test is applied once to the whole dataset after the validation process.

Another important modification is the addition of unbiased feature importance through random permutation of OOB observations. We believe that this property is a cornerstone of the proposed approach and inline with the current clinicians’ needs. While they need to know if two groups are (or are not) significantly separated, they are also interested to know the most influential features that lead to the reported result. Although the algorithm offers this convenience, we need to note that feature importance should be treated with extra care. The proposed approach tries to minimize the false conclusions concerning the importance of features when redundant or highly colinear features are present but the above topic is still under research. A general advice to clinicians can be to check for features exhibiting mutual information before the beginning of the testing process.
  
\subsection{Population effect}  

The features computed by the basic Romberg test have been reported as relatively inconclusive in distinguishing fallers and non-fallers, mainly due to the lack of realistic conditions of fall \cite{palmieri2002center}. The available patients' dataset, with its relatively "marginal" separation between fallers and non-fallers (see Tab.~\ref{Table 3.}), can be considered as an ideal dataset in order to check the performance of the newly proposed approach. We consider MMD algorithm (see Sec.~\ref{sec:experimental_setting}) as the gold-standard method in terms of separability of the two groups. The fact that ts-AUC shows similar performance to that of MMD is very important, especially if we think that the proposed ts-AUC can also provide additional information about the most influential features without the need of any supplementary (meta-)analysis. Therefore, it would be fare to say that ts-AUC is competitive in terms of performance, while also boosting the interpretability of the result for the convenience of clinicians. 

Interestingly, the decrease of the overall population and the gradual balancing between the groups of fallers and non-faller, showed that the proposed test is less conservative than the multiple testing process (with corrections). Exploratory studies, where a hypothesis about the structure of the dataset is not strictly defined in advance, could benefit from such multivariate approaches.  

Comparing the results of the two population reduction schemes, i.e.~the uniform  reduction of the population versus the reduction of non-fallers (the larger group), we observe that all the statistical tests performed slightly worse in the former case. This was an expected result since fallers were only 24 out of the 123 available PS patients, and thus decreasing the size of that group made the fallers heavily underrepresented in the produced subsample. 

\subsection{Limitations}  

The first limitation of this study is the lack of sufficient evidence about the reasons behind falls. The basic Romberg test has been reported to be an insufficient protocol to provide such physiological information \cite{palmieri2002center,swanenburg2010falls}. Previous studies proposed richer protocols (including multi-tasking or use of foam surfaces \cite{Melzer2004,chagdes2009multiple,swanenburg2010falls}) for postural control assessment of fragile individuals such as PS patients. Undoubtedly, such protocols can have beneficial effect to the faller/non-faller classification, as well as to the impairment assessment of patients (visual, vestibular, somatosensor, nervous system). Yet, among the objectives of this work was to show that basic Romberg test does contain fall risk-related information, whose extraction and full exploitation is largely up to the adequacy of the employed statistical analytics. 

It is worth noting that there is always some uncertainty in what patients report as their recent fall experience. Participants who were asked about previous falls might confabulate without a conscious intention to deceive (recall bias). Therefore, some of the non-fallers might be mistakenly labeled as non-fallers. Machine learning algorithms are usually robust to the presence of such noise which in our opinion is always minor.  

In extreme cases of imbalanced datasets with many negative values and few positive ones, other metrics rather than AUC, such as precision-recall (PR) curve, F$_1$ score or area under the PR curve, could be more appropriate in order to control possible overfitting \cite{davis2006relationship}. We decided to keep the criterion (AUC), initially proposed by \cite{Vayatis2009} for balanced datasets, in order to fulfill one of our main objectives: to propose the algorithm as  understandable, interpretable and easy-to-implement as possible. In return, as it has been already mentioned, we controlled the leaf size ($LS$) and features' number ($M$) optimization procedure, and we applied cross-validation in each resulting case.

The use of Wii Balance Board (WBB) as a force platform during the acquisition protocol, is another mentionable limitation. The reliability of the WBB as a medical examination tool has been previously questioned \cite{pagnacco2011biomedical}. Basic reported drawbacks were: a)~the modest agreement with laboratory grade force platforms, b)~the lower signal to noise ratio in its recording, and c)~the irregular sampling rate \cite{castelli2015we}. We state that we are perfectly aware of the aforementioned limitations. However, the WBB presents an increasing popularity in posturography studies  as a valid tool for assessing standing balance \cite{clark2010validity,leach2014validating}. It is an inexpensive piece of equipment and hence seems ideal for applications that intend to provide a quick and low-cost first scan of individuals with certain possibility of postural control loss. In addition, recent works \cite{leach2014validating,audiffren2016preprocessing} showed that a careful preprocessing can mitigate some of its aforementioned drawbacks.

\section{Conclusions and perspectives}\label{sec:conclusions}

In this paper we showed that using the proposed ts-AUC test, which is a two-sample test based on AUC maximization, faller and non-faller patients who suffer from Parkinsonian syndromes (PS) can actually be distinguished by examining posturographic features that are derived following the basic Romberg protocol. This novel approach was also able to indicate the posturographic features that are significantly different between the two groups. We confirmed that a fall-prone PS patient may manifest wider and more abrupt antero-posterior oscillations and larger posturographic areas compared to a non-faller. This separation appeared statistically less detectable when using more traditional approaches such as multiple testing. Interestingly, the above results were observed only in statokinesigrams derived by the Eyes-open protocol. The results of our study have highlighted that new multivariate methods based on machine learning, such as ts-AUC, can play an important role in assessing the usefulness of simple and inexpensive acquisition protocols as well as the extracted posturographic features.

\section*{Acknowledgment}

The authors would like to thank Julien Audiffren for the initial database construction and the implementation of the statokinesigrams' preprocessing (SWARII algorithm \cite{audiffren2016preprocessing}) that we have used. We also thank Albane Moreau for providing the additional database information concerning the PS patients. Part of this work was funded by the IdAML Chair hosted at ENS-Paris-Saclay.

\bibliographystyle{IEEEbib}

\bibliography{TwoSamplePosturography}

\end{document}